# Osteoarthritis Disease Detection System using Self Organizing Maps Method based on Ossa Manus X-Ray

Putri Kurniasih
Department of Information Engineering of Trisakti University, Faculty of Industrial Technology

Dian Pratiwi
Department of Information Engineering of Trisakti University, Faculty of Industrial Technology

## ABSTRACT
Osteoarthritis is a disease found in the world, including in Indonesia. The purpose of this study was to detect the disease Osteoarthritis using Self Organizing mapping (SOM), and to know the procedure of artificial intelligence on the methods of Self Organizing Mapping (SOM). In this system, there are several stages to preserve to detect disease Osteoarthritis using Self Organizing maps is the result of photographic images ronsen Ossa manus normal and sick with the resolution (150 × 200 pixels) do the repair phase contrast, the Grayscale, thresholding process, Histogram of process , and do the last process, where the process of doing training (Training) and testing on images that have kept the shape data (.text). the conclusion is the result of testing by using a data image, where 42 of data have 12 Normal image data and image data 30 sick. On the results of the process of training data there are 8 X-ray image revealed normal right and 19 data x-ray image of pain expressed is correct. Then the accuracy on the process of training was 96.42%, and in the process of testing normal true image 4 obtained revealed Normal, 9 data pain stated true pain and 1 data imagery hurts stated incorrectly, then the accuracy gained from the results of testing are 92,8%.

## 1. INTRODUCTION
Osteoarthritis is a disease of the joints are the most common case in the world. [1] The prevalence of osteoarthritis in Indonesia 34.3 million people in 2002 and reached 36.5 million people in 2007. An estimated 40% of the population aged above 70 years suffering from osteoarthritis, and 80% of patients osteoarthritis has the limitation of motion in varying degrees of light until the weight of the resulting reduced the quality of her life because the pretty high prevalence. [2] The image is a two-dimensional function f (x, y) that States the intensity of light, where x and y the spatial coordinates and f stated at point (x, y) declaring the brightness (brightness) image at that point. The image is being processed and analyzed in the form of a photo or image that has been diskritkan into a digital image. The value of viualisasi determines the intensity up to black color white declared as grey level or degree to Grey's. Severity of Disease Detection System of Osteoarthritis in the Manus method using Self Organizing Maps, This detection system which aims to see the severity of the disease arthiritis and know the procedures of artificial intelligence Neural Network on the methods of Self Organizing Mapping (SOM).  A The reason to use the method of Self Organizing Maps (SOM). Neural network is a competitive and cooperative that can accommodate the input vectors of topology structure and maps on a high-dimensional network data to lower, then form the map of topology. Neural network have good results to  implement in the medical field. As in a study conducted by previous researchers who used Perceptron Neural Network method in dengue fever case, has resulted in a maximum accuracy of pattern recognition of 80% [10]. So that the researcher in this study still use neural network method.

## 2. THE STUDY OF LIBRARY
### 2.1  Ossa Manus
Understanding the limitations of the hand (*ossa manus*) is a hand, the distal region is of upper limb, including the carpus (*ossa carpalia, wrist*), metacarpus, and digiti (*ossa phalanges*). Ossa manus, based on anatomical position, from proximal to distal, consisting of carpal, and phalanges. [3] The following image below explains about the bones of the hand :

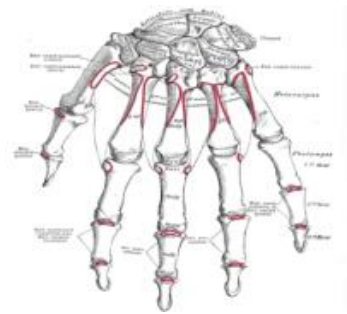

**Figure 1** *Ossa manus*

### 2.2  Introduction of  Disease Osteoarthritis

*Osteoarthritis* (OA) is a degenerative joint disease, which the entire structure of the joints experience pathological changes. Characterized by demage to the cartilage hyalin joints, increasing the thickness as well as multiple sclerosis from plates of bone, the growth of osteophytes on the shores of the joints,stretched his kapsula joints, the onset of inflammation, and the weakening of muscles-muscles that connect the joints. [4]

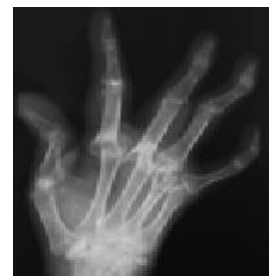

**Figure 2** *Osteoarthritis pada Ossa Manus*

Based on the cause, Osteoarthritis is distinguished into two primary and secondary Osteoarthritis. Primary osteoarthritis or osteoarthritis can be called idiopathic, has no definite cause (unknown) and not caused by systemic disease and the process of local changes in the joints. Secondary osteoarthritis, in contrast to primary Osteoarthritis, Osteoarthritis is caused by inflammation, endocrine system disorders, metabolic, growth, heredity (hereditary), and prolonged immobilization. In General, patients Osteoarthritis said that complaints that he felt had been longstanding, but growing slowly. Here are the complaints that can be found on the patients of Osteoarthritis: Joint pain, barriers to movement of the joints, Enlarged joints (deformities), swelling of the joints that are asymmetrical, the sign – the sign of inflammation.[2]

## 2.3 Digital Image Processing
The digital image is an image that has been saved in the file so that it can be processed with the use of a computer. A digital image is a two dimensional array or a matrix the elements stated level of keabuan elements of the image. The digital image is not always a direct result of the data recording system. Sometimes the result data records are continuous as the images on television monitors, x-rays, etc. The basic elements of the digital image is at the brightness (brightness), contrast (contrast), contours (contour), color (colour), form (shape), texture (texture). [5]

## 2.4 Image Quality Improvement
Improvement of the quality of the image (Image enhancement) was one of the early processes in image processing (image processing). The process of setting the contrast value of the multiplication process is the degree of keabuan change the value of x with the contrast $t_{contrast}$.

$$x_{contrast} = x \times t_{contrast} \qquad (1)$$

$0 < t_{kontract} < m$, with positive m

Improvement of quality is necessary because often times the image was made the object of deliberations have poor quality, for example the image undergone noise (noise) upon delivery through the transmission line, the image is too light or dark, blurred image less sharp, and so on. The processes included in the image quality improvement that is Changing the brightness of the image (image brightness), Stretching contrast (contrast strerching), Histogram Equalization, Smoothing image (image smoothing), Refine (sharpening) edge (edge), and Colour apparent (pseudocolouring ). [5]

## 2.5 Grayscale
Grayscale image is grayscale color image using the color depth of the color grayShades of gray is the only color in RGB space with components of red, green, and blue have the same intensity.[6] Grayscale image is stored in 8-bit format for each sampled pixel, allowing as many as 256 intensity. Changing the color image that has a value matrix, each R, G and B into a grayscale image with a value of X, then the conversion can be done by taking the average of the values of R, G and B. To get a grayscale image (graying) used the formula:

$$I(x,y) = R(x,y) + G(x,y) + B(x,y)/3 \qquad (2)$$

*Where: fr is the value component of red, fg is the value component of the green, and blue component is the value of fb.*

## 2.6 Thresholding
Thresholding (threshold) is the process of splitting pixels-pixels based on the grey level, which can be used to split areas between background and foreground [11]. Pixels that have a degree of gray is smaller than the specified limit value shall be given a value of 0, while pixels that have a degree of gray is greater than the limit value will be changed to 1. Thresholding is used to set the number of degrees of gray on the image. By using thresholding then degrees of gray can be changed as you wish. This thresholding process is basically a process of conversion quantization on the image, so as to perform the thresholding with degree of gray can use the formula: [7]

$$T = \frac{1}{2}(\mu_1 + \mu_2) \qquad (3)$$

## 2.7 Histogram And Normalization Of Data

### 2.7.1 Histogram
The image histogram is a graph that can be used to find out the distribution of levels of gray an image.[8] of the Histogram can be calculated with the following formula:

$$h_i = \frac{n_i}{n} \qquad (4)$$

*Where: $n_i$ is the number* of *pixels that have a value of gray I (i = 0 ... ... L-1), L is the maximum interval color, n is the total of the pixels in the image, $h_i$ is the probability of the value of gray i.*

### 2.7.2 Normalization Of Data
Normalizing data is a method for grouping a range or interval of different values to the same scale. Normalizing essential used to give equal weight against the values of the different features of the results of the extraction. Normalization on vector feature can be done in various ways, one of which, by using methods of the Min – Max Normalization:

$$D'(i) = \frac{D(i) - min(D)}{max(D) - min(D)} \times (U - L) \qquad (5)$$

Where: *D'* is the result of data normalization, *D* is the value before normalization, *U* is the upper limit value (*Upper bound*), *L* is the value of the upper limit (*Lower bound*).

## 2.8 Self Organizing Maps
Self-organizing Folder (SOM) was first introduced by Kohonen training techniques with ANN, which uses a base of winner takes all, where only the winner neuron will be updated does it weigh. For learning algorithms one of SOM is the algorithm.

The algorithm on the SOM neural network as follows [4] :

a. If the feature vector matrix of size k x m (k is the number of feature vector dimensions, and m is the number of data), the initialization :

- The number of the desired j class or cluster

- The number of component i of the feature vector matrix (k is the row of matrix)

- The number of vector Xm,i = amount of data(matrix column)

- The initial weights Wji were randomly with interval 0 to 1

- The initial learning rate α(0)
- The number of iteration (e epoch)

b. Execute the first iteration until the total iteration (epoch)

c. Calculate the vector image to start from 1 to m :

For all of j

- Then determine the minimum value of D(j)
- Make changes to the j weight with the minimum of D(j)

d. Modify the learning rate for the next iteration :

$$\alpha(t + 1) = 0,5 \, \alpha(t)$$

*which t start from the first iteration to e.*

e. Test the termination condition

Iteration is stopped if the difference between Wji and Wji the previous iteration only a little or a change in weights just very small changes, then the iteration has reached convergence so that it can be stopped.

f. Use a weight of Wji that has been convergence to grouping feature vector for each image, by calculating the distance vector with optimal weights.

g. Divide the image (Xm) into classes :

1. If $D(1) < D(2)$, then the images included in class Normal
2. If $D(2) < D(1)$, then the images included in class Sick

## 3. FIGURES/CAPTIONS

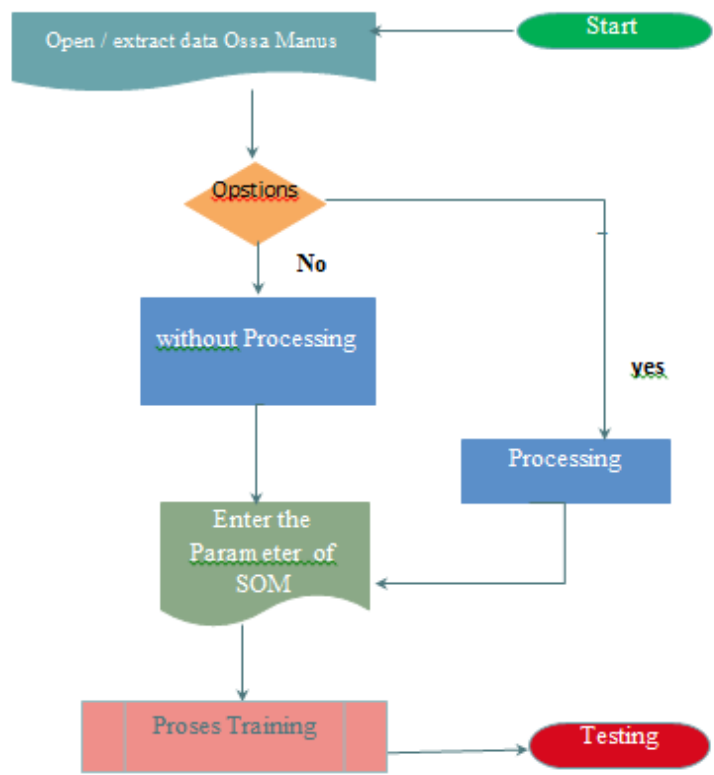

**Figure 3** Flow Chart of Osteoarthritis disease Detection System

## 4. RESULTS AND DISCUSSIONS
### 4.1 Menu File

1. *Open Image*

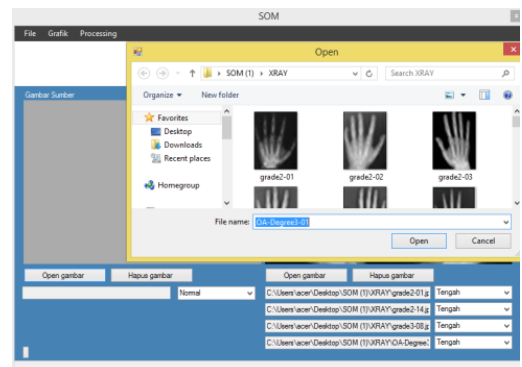

**Figure 4** *Open Image*

In Figure 4.7, the process of the Open Image or opening an image is the first step to start a process. There are 2 images that will open IE the source image (photo ronsen normal) and the image will be tested (photos ronsen is not normal). The source image will open in a normal x-ray folders, while the image you want in the test will be in the open in the folder x-ray.

2. *Save Image*

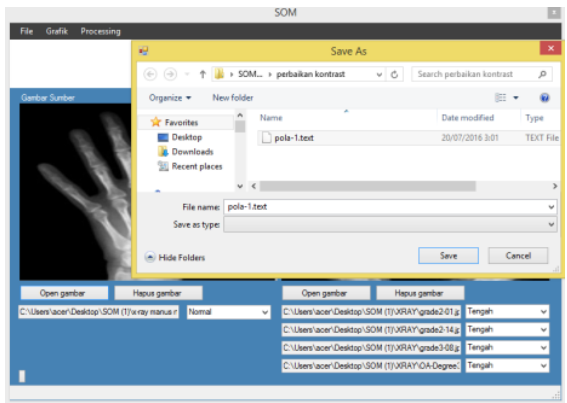

**Figure 5** *Save the image on the File menuExit*

In Figure 2.9, process or Save Image to save the picture is a picture that was already in the process of a number of processes will be saved in the folder of each. The image will be saved in the form of text (text).

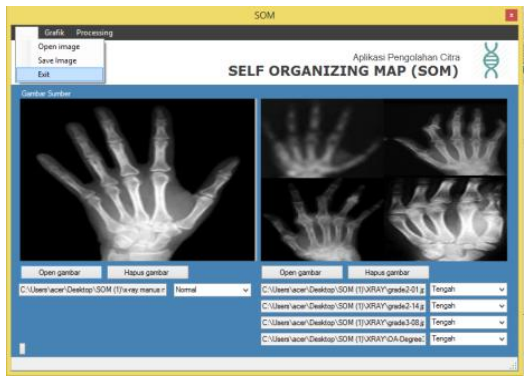

**Figure 6** *Exit from the program when the process runs*

In Figure 2.9, this Exit button will work when we want to terminate the process when the process is running.

## 4.2 Display Contrast Improvement

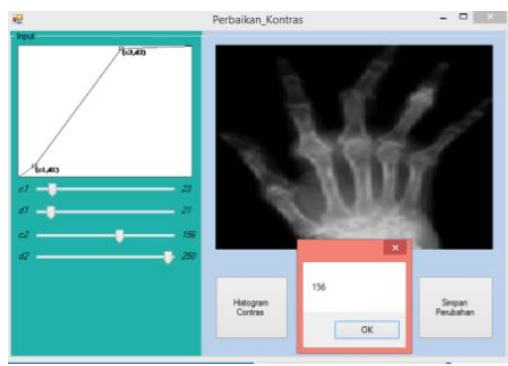

**Figure 7** *The results contrast improvement process*

In Figure 4.8, image contrast improvement process using contrast stretching is not much effect on 55% of the processed image if it is done using with auto rates menu. The results of the image does not always seem to be better seen from the quality of the color.

## 4.3 Threshold Display

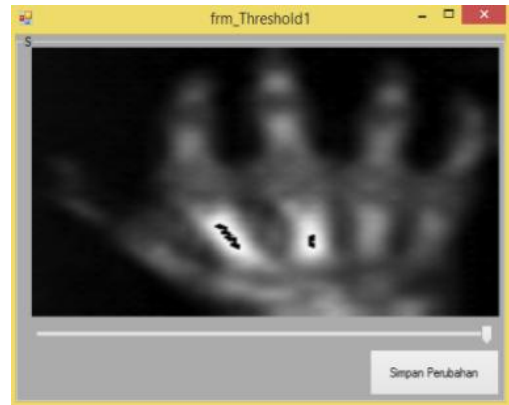

**Figure 8** *The results in Figure threshould pain*

In Figure 4.11, the next step is to change the image thresholding image into binary so that this process of background of the image can be removed, or the value of the background change menjdi 0 and the value of tulnag is not changed.

## 4.4 Histogram Display

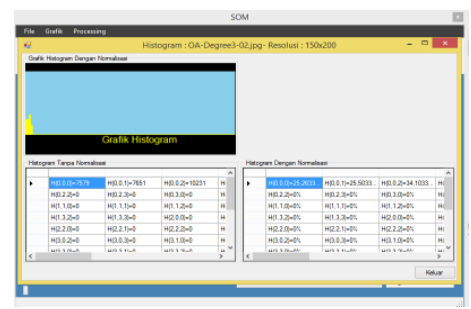

**Figure 9** *The results of the process of Histogram*

In Figure 4.12 is page view histogram. Here after doing some process then we wont see results in the form of value. Histogram data is divided into two, namely: Histogram without Normalization and Histogram with Normalization. Here is the Histogram with Normalization is useful to look at the whole range of grayscale values so obtained at sharper image, compared to data Histogram without normalization.

## 4.5 *Self Organizing Maps* Display

In Figure 4.15, this stage is done after stage dimension reduction is the process of training or training, namely by applying the method of SOM on data training. Where the trial results in the process of training can evaluate the cluster results in doing with how to calculate: Sum Squared Error (SSE) is useful as a measure of the variation in a cluster, Avarage Quantization Error (AVG) useful mengavaluasi or calculate the average quantization error using data that has been in training, data Normalization, the value of Grayscale, Area.

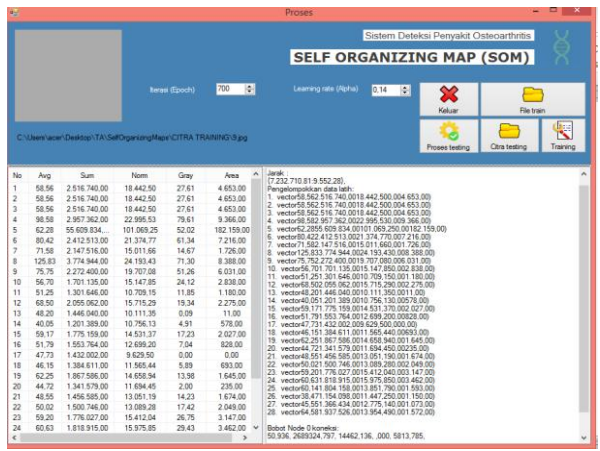

**Figure 10** *The results of the process of Training from x-ray image data*

After the maximum iteration reached, obtained the final weights of the most constant. The weights that are later used to process trials against data testing.

On the chart Table 4.10 there are 42 x-ray image data based on the diagnosis of doctors who will be in training (Training): 12 X-ray image data normally and 30 x-ray image data pain. On the results of the process of training data there are 8 X-ray image of a normal, true stated normal and 9 x-ray image data correctly expressed pain and 1 x-ray image data was declared incorrectly. Then the results of the curatorial process training was 96.42%.

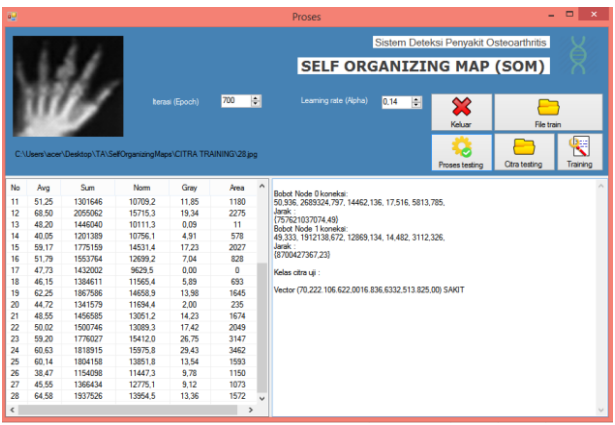

**Figure 12** *The results of the process of Testing on the test image data*

In Figure 4.17, the process of testing no longer apply methods SOM, but only by measuring the distance between the data testing with weights. This process is called Euclidean distance or by calculating the Euclidean distance. After a note on weighted nth smallest distance is obtained against the testing data to-m, then the results of the cluster shows the data to-m included in cluster n.

**Tabel 1** *The results of the process of Testing on the test image data*

| No. | Figure normal and diseased | Diagnosis Doctors | Testing |
|---|---|---|---|
| 1. | Data 1 | Healthy | Normal |
| 2. | Data 2 | Healthy | Normal |
| 3. | Data 3 | Healthy | Normal |
| 4. | Data 4 | Healthy | Normal |
| 5. | Data 5 | Sick | Sick |
| 6. | Data 6 | Sick | Sick |
| 7. | Data 7 | Sick | Sick |
| 8. | Data 8 | Sick | Sick |
| 9. | Data 9 | Sick | Sick |
| 10. | Data 10 | Sick | Sick |
| 11. | Data 11 | Sick | Sick |
| 12. | Data 12 | Sick | Sick |
| 13. | Data 13 | Sick | Sick |
| 14. | Data 14 | Sick | Normal |

According to Table 4.3 testing Results obtained from 2 X-ray image data, image data 10: sick and 4 normal image data. Where the normal image 4, acquired right is declared Normal, 9 data pain expressed is true ill and sick image data revealed 1 wrong. Then the accuracy of the test results obtained from testing is 92,8571%.

## 5. CONCLUSIONS

1. Detection system of this disease can determine the severity of osteoarthritis by using *Self Organizing maps* (SOM) method. The reason, because this algorithm has been successfully developed to detect *Osteoarthiritis* diseases using *Self Organizing Maps* method on image of X-Ray manus.

2. The cluster formation of this detection system using SOM is by training process with parameter: iteration is 700, and learning rate is 0.1, then the data yields the cluster number, w weights (i, j), and the distance weight on the cluster , Then the data in the process into the testing (testing) and issued the final result (output) that is normal or sick class.

3. Self Organizing Maps test results using 42 image data, where 12 Normal image data and 30 image data Pain. In the result of training process according to doctor's prediction there are 8 normal X-ray image data and 20 X-ray image data is sick, and in the testing process obtained 3 normal image correctly stated Normal, 10 sick data stated true ill and 1 Normal image data stated wrong. Then the accuracy obtained from test testing result is 92,85%.